\documentclass[10pt,twocolumn,letterpaper]{article}

\usepackage{wacv}              

\usepackage{graphicx}
\usepackage{amsmath}
\usepackage{amssymb}
\usepackage{booktabs}
\usepackage{graphicx}
\usepackage{latexsym}
\usepackage{amsmath}
\usepackage{algorithm}
\usepackage{algpseudocode}
\usepackage{xcolor}
\usepackage{mathalfa}
\usepackage{amssymb}
\usepackage{multirow}

\makeatletter
\robustify\@latex@warning@no@line
\makeatother
\usepackage[noblocks]{authblk}
\usepackage[title]{appendix}

\makeatletter
\renewcommand\AB@affilsepx{, \protect\Affilfont}
\makeatother

\usepackage[pagebackref,breaklinks,colorlinks]{hyperref}

\usepackage[capitalize]{cleveref}
\crefname{section}{Sec.}{Secs.}
\Crefname{section}{Section}{Sections}
\Crefname{table}{Table}{Tables}
\crefname{table}{Tab.}{Tabs.}

\def\wacvPaperID{2044} 
\def\confName{WACV}
\def\confYear{2024}

\begin{document}

\title{Looking through the past: better knowledge retention\\ for  generative replay in continual learning}

\author[1]{Valeriya Khan \thanks{corresponding author, email: \href{mailto:valeriya.khan@ideas-ncbr.pl}{valeriya.khan@ideas-ncbr.pl}}}
\author[1,2]{Sebastian Cygert}
\author[1,3]{Kamil Deja}
\author[1,3,4,5]{Tomasz Trzciński}
\author[2,6,7]{Bartłomiej Twardowski}
\affil[1]{\normalsize IDEAS NCBR}
\affil[2]{Gdańsk University of Technology}
\affil[3]{Warsaw University of Technology}
\affil[4]{Jagiellonian University}
\affil[5]{Tooploox}
\affil[6]{Computer Vision Center}
\affil[7]{Universitat Autònoma Barcelona}
\maketitle
\begin{abstract}
In this work, we improve the generative replay in a continual learning setting to perform well on challenging scenarios. Current generative rehearsal methods are usually benchmarked on small and simple datasets as they are not powerful enough to generate more complex data with a greater number of classes. We notice that in VAE-based generative replay, this could be attributed to the fact that the generated features are far from the original ones when mapped to the latent space. Therefore, we propose three modifications that allow the model to learn and generate complex data. More specifically, we incorporate the distillation in latent space between the current and previous models to reduce feature drift. Additionally, a latent matching for the reconstruction and original data is proposed to improve generated features alignment. Further, based on the observation that the reconstructions are better for preserving knowledge, we add the cycling of generations through the previously trained model to make them closer to the original data. Our method outperforms other generative replay methods in various scenarios. Code available at \href{https://github.com/valeriya-khan/looking-through-the-past}{https://github.com/valeriya-khan/looking-through-the-past}.
\end{abstract}

\section{Introduction}
\label{sec:intro}

The traditional approach to machine learning involves training models on shuffled training data to ensure independent and identically distributed conditions, enabling the model to learn generalized parameters for the entire data distribution. On the other hand, in continual learning, the models are trained on sequential tasks, with only data from the current task available at any given time. Such scenario is more realistic in some applications with, for example, privacy concerns, where the old data may become unavailable. However, models trained in such an incremental fashion will face a catastrophic forgetting~\cite{Mccloskey89}, a significant drop in the accuracy of previously acquired knowledge.   

A popular setting for continual learning is Class Incremental Learning (CIL), where the goal is to train the classifier on new classes in consequent incremental steps~\cite{Masana2022}. Typically, different types of regularizations are applied~\cite{li2016_lwf,zenke2017si}, however, without using any exemplars of the previous tasks, the results are far away from being satisfactory. Hence, there is an interest in generative models~\cite{goodfellow2014gans}, which allow replaying the synthetic data from previous tasks using a trained generative model.  
 
Despite the promising setup, it turns out to be very challenging to scale approaches based on generative models in CIL to more demanding datasets than MNIST or CIFAR-10~\cite{shin2017deepgenerativereplay}. Generative replay models often have poor results on datasets with more complex data or a greater number of different classes~\cite{lesort2019generative}. 
This is mainly because modeling high-dimensional images in incrementally trained generative models is very challenging, as from task to task the quality of generated data degrades.
Therefore, some recent works~\cite{liu2020generative} incorporated feature-based replay when the data is first passed through the trained and frozen feature extractor, and only then it is used for training the generator part. One significant benefit of utilizing feature replay is that the distribution that needs to be learned by the generative model is usually much simpler and has lower dimensionality.

One of the recent works in the generative replay that utilizes the feature replay is Brain-Inspired Replay (BIR)~\cite{van2020brain}. In their work, the authors introduce several modifications to make variational autoencoder able to learn and generate more complex data, even in long sequences. 
The highest results reported by the authors are when BIR is combined with Synaptic Intelligence (SI)~\cite{zenke2017si} regularization method, which suggests that BIR alone for a generative features-replay is not enough and maybe other regularization techniques can yield better results. It motivates us to analyze an in-depth VAE-based replay approaches with BIR as its flagship example.
We observe, that there is still a significant difference between features from the real data and those produced by the generator. We hypothesize that this may have a detrimental effect on the quality of the data replay, and hence we add two modifications to the model that mitigate the problem.
Firstly, we introduce a new loss term for minimizing the difference between the encoded latent vectors of the original sample and the reconstructed sample. This loss enables the encoder to learn how to reverse the operation of the decoder. 
Secondly, we propose to refine the quality of rehearsal samples. To that end, we introduce a cycling method where we iterate the generated data through the previously trained model (decoder and encoder), and only after that feed it to the replay buffer for training the new model. As we show in our analysis, this has the effect of
reducing a discrepancy between original and generated features for a classification (see Figure \ref{teaser}), and as a result, improves the final model accuracy. The proposed changes allowed us to significantly improve the results over our baseline method.

\vspace{1em}
Overall, the main contributions of this work are threefold:
\begin{figure*}[ht]
\centerline{\includegraphics[width=1\linewidth]{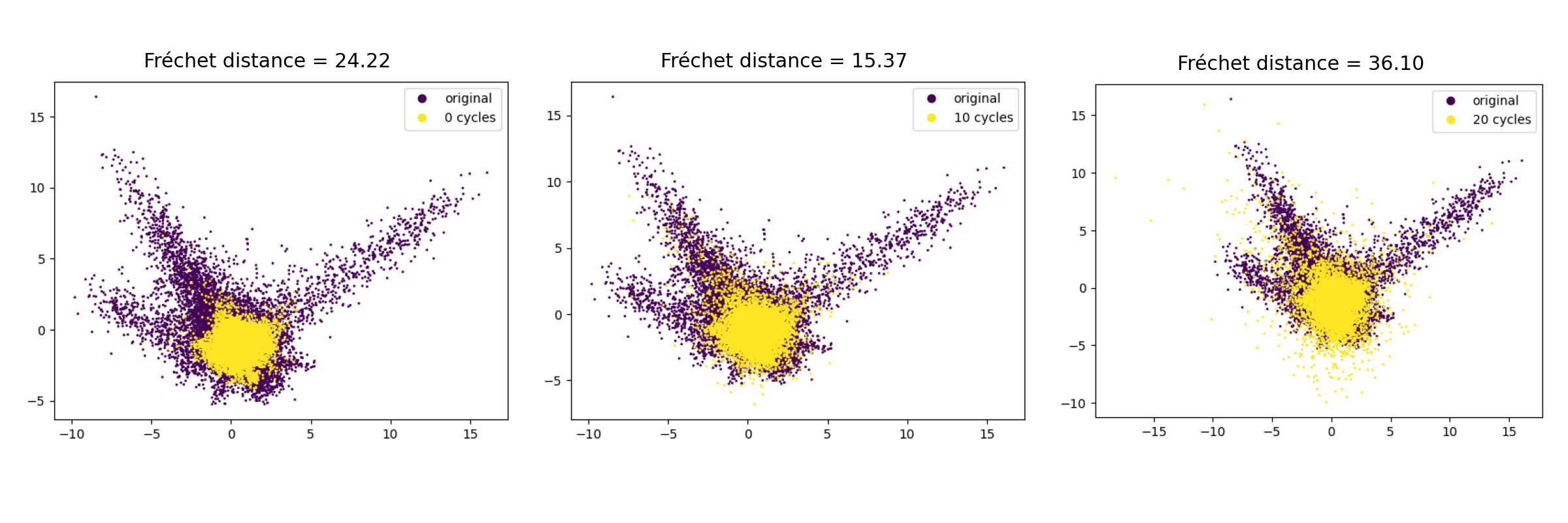}}
\vspace{-1cm}
\caption{Principal Component Analysis (PCA) plots were computed on original latent vectors and generated ones when doing 0, 10, and 20 cycles respectively. By looking at both the PCA plots and Fr\'echet distances we can observe the generated latents are more aligned with the original ones when using an appropriate number of cycles.} \label{teaser}
\end{figure*}

\begin{itemize}
\item We analyze existing feature-generative replay methods for class-incremental learning and identify the weaknesses of recent VAE-based approaches, such as degraded generated samples and a mismatch in the distribution of current (original) features and old (generated) ones.
\item Building on our analysis, we propose a new method for class-incremental learning with generative feature replay. Our method improves the matching of latent representations between reconstructed and original features through distillation, and generations' cycling to effectively reduce the discrepancy between new and old samples for classification.
\item Through a series of experiments, we demonstrate that our method significantly outperforms the baseline approach (BIR), without requiring additional SI~\cite{zenke2017si} regularization. Furthermore, our ablation study shows that each introduced modification contributes incrementally to the overall improvement in the model's accuracy.
\end{itemize}
\section{Related works}

Continual learning methods can be divided into three categories that we overview in this section.

\emph{\textbf{Regularization methods}} aim to strike a balance between preserving previously acquired knowledge and providing sufficient flexibility to incorporate new information. To that end, regularisation is applied to slow down the updates on the most important weights. In particular, in Elastic Weights Consolidation (EWC)~\cite{kirkpatrick2017overcoming} authors propose to use Fisher Information to select important model's weights, while in Synaptic Intelligence (SI)~\cite{zenke2017continual} and Memory Aware Synapses (MAS)~\cite{aljundi2018memory} additional information is stored together with each parameter. Similarly, in Learning Without Forgetting (LWF)~\cite{li2017learning} additional distillation loss on current data is used to match the output of the model trained on the previous task, with a new one. In this work, we use distillation techniques to align representations of old and new features similarly to LWF.

\emph{\textbf{Dynamic architecture}} methods create different versions of the base model for each task. This is usually implemented by creating additional task-specific submodules~\cite{2016rusu+7, 2017yoon+3, 2018xu+1}, or by selecting different parts of the base network~\cite{2018masse+2,2018mallya+1,2019golkar+2,2018mallya+2}. Such approaches reduce catastrophic forgetting at the expense of expanding memory requirements. 

\emph{\textbf{Rehearsal methods}} involve storing and replaying past data to prevent catastrophic forgetting. The simplest implementation of this approach employs a memory buffer where a subset of examples from previous tasks can be stored~\cite {isele2018selective,lopez2017gradient,2019chaudry+6,belouadah2019il2m,prabhu2020gdumb}. Such an approach achieves high performance and can significantly reduce catastrophic forgetting. 

However, the memory buffer has to store a significant number of examples and, hence, grow with each task. Also in some domains, due to privacy concerns, using historical data is not possible. Therefore, generative models are often used to synthesize past data. The first example of \emph{\textbf{generative replay}} for CIL model is~\cite{shin2017deepgenerativereplay} where a generative model (e.g., Generative Adversarial Network (GAN)~\cite{goodfellow2014gans}) is used as a source of rehearsal examples. This idea is further extended to other generative methods such as Variational Autoencoders~\cite{kingma2014autoencoding} in \cite{van2018generative, 2020mundt+4} or Normalising Flows~\cite{rezende2015variational} in \cite{scardapane2020pseudo}. In \cite{lesort2019generative}, the authors overview the general performance of generative models as a source of rehearsal examples, showing that even though GANs outperform other solutions, all the methods struggle when evaluated on more complex benchmark scenarios. Therefore, to simplify the problem, in Brain-Inspired Replay (BIR)~\cite{van2020brain} the authors introduce a new idea known as \emph{feature replay} and propose to focus on the replay of internal data representations instead of the original samples. This idea was further explored in~\cite{kemker2017fearnet}, with a split between short and long-term memory, and in~\cite{liu2020generative} where authors employ conditional GANs. Our method presented in this work falls in the generative-feature replay category, as we directly base our approach on the BIR method.

\section{Method}

\subsection{Problem definition}

In this work, we focus on image classification in a class-incremental setting. The model is trained on the sequence of tasks $T_1, T_2, ..., T_n$. The training data $\{X^{(t)}, Y^{(t)}\}$ is drawn from the distribution $D^{(t)}$, where $X^{(t)}$ are the training samples, $Y^{(t)}$ are the ground truth labels, and $1\leq t\leq n$ is the current task id. In this context, the \emph{task} means an isolated training phase with access only to this task data (cannot recall old data). 
As we consider class-incremental learning, where the model has to be trained to predict the labels for all of the tasks seen so far. 

\subsection{Baseline model}
\label{sec:base}
Our work is based on the Brain-Inspired Replay (BIR) method~\cite{van2020brain}. The model contains two main parts: a pre-trained feature extractor and a symmetrical VAE on top of it. The VAE is used as a feature generator in BIR to replay old knowledge. It consists of the encoder $q_{\phi}$ and the decoder $p_{\psi}$. The encoder maps the input $x$ to stochastic latent variables $z$, and the decoder maps these latent variables back to reconstructed vector $\hat{x}$.
Usually, a VAE model is trained by maximizing the evidence lower bound (ELBO), which is analogous to minimizing the following per-sample loss:
\begin{multline}
    L^{G}(x;\phi,\psi) = E_{z\sim q_{\phi}(.|x)}[-\log p_{\psi}(x|z)]+\\ + D_{KL}(q_\phi(.|x)||p(.))\\
=L^{recon}(x;\phi,\psi)+L^{latent}(x;\phi),
\end{multline}

where 
\begin{math}
    q_{\phi}(.|x) = \mathcal{N}(\mu^{(x)},\sigma^{(x)^2}I)
\end{math}
and
\begin{math}
    p(.)=\mathcal{N}(0,I)
\end{math}
are the posterior and prior distributions over the latent variables respectively, and $D_{KL}$ is the Kullback-Leibler divergence.
For prior distribution equal to $N(0,I)$, the KL divergence can be calculated as follows:
\begin{equation}
    L^{latent}(x;\phi) = \frac{1}{2} \sum_{j=1}^{D}(1+\log (\sigma_{j}^{(x)^2})-\mu_j^{(x)^2}-\sigma_j^{(x)^2}),
\end{equation}
where $D$ is a latent dimension. The reconstruction loss in this work is given by:
\begin{multline}
        L^{recon}(x;\phi,\psi) = E_{\epsilon\sim \mathcal{N}(0,I)}\bigg[\sum_{p=1}^{N}x_p\log (\hat{x}_p)\\ +(1-x_p)\log(1-\hat{x}_p) \bigg],
\end{multline}
where $N$ is the size of the input, $x_p$ is the $p^{\text{th}}$ entry of the original input $x$, and $\hat{x}_p$ is the $p^{\text{th}}$ entry of reconstruction $\hat{x}$.

To generate samples of specific classes, the standard normal prior is substituted by the Gaussian mixture with a separate distribution for each class:
\begin{equation}
    p_{{}_\mathcal{X}}(.) = \sum_{c=1}^{N_{\text{classes}}}p(\mathcal{Y}=c)p_{{}_\mathcal{X}}(.|c),
\end{equation}
where $p_{{}_\mathcal{X}}(.|c) = \mathcal{N}(\mu^c,\sigma^cI)$ for $c=1, ..., N_{\text{classes}}$, $\mu^c$ and $\sigma^c$ are trainable means and standard deviation for class $c$, $\mathcal{X}$ is a set of means and standard deviations for all classes $N_classes$ and $p(\mathcal{Y}=c)$ is the class prior.

For the current task with hard targets (labels), the $L^{latent}$ has the following form:
\begin{multline}
        L^{latent}(x,y;\phi, \mathcal{X}) = \frac{1}{2} \sum_{j=1}^{D}\bigg(1+\log (\sigma_{j}^{(x)^2})\\-\log(\sigma_{j}^{y^2})-\frac{(\mu_j^{(x)}-\mu_j^{y})^2+\sigma_{j}^{(x)^2}}{\sigma_{j}^{y^2}}\bigg),
\end{multline}
where $\mu_j^{y}$ is the  $j^{\text{th}}$ element of $\mu^{y}$ and $\sigma_{j}^{y}$ is the  $j^{\text{th}}$ element of $\sigma^{y}$.
For the replay, this loss is estimated for soft-target $\Tilde{y}$ as:
\begin{multline}
     L^{latent}(x,y;\phi, \mathcal{X}) = \frac{1}{2} \sum_{j=1}^{D}\bigg(1+\log(2\pi)+\log(\sigma_{j}^{(x)^2}) \bigg)\\
     +E_{\epsilon\sim \mathcal{N}(0,I)}\Bigg[\log \bigg(\sum_{j=1}^{D}\Tilde{y}_j\mathcal{N}(\mu^{(x)}+\sigma^{(x)}\odot\epsilon|\mu^j,\sigma^{j^2}I) \bigg) \Bigg],
\end{multline}
where $\Tilde{y}_j$ is the $j^{\text{th}}$ entry of $\Tilde{y}$, and estimation of expectation is performed by a single Monte Carlo sample for each input. 

For the current task, classification loss is given by:
\begin{equation}
    L^C(x,y;\theta) = -\log p_\theta (\mathcal{Y}=y|x),
\end{equation}
where $p_\theta$ is the conditional probability distribution defined by the parameters of the model.

For the replay part in BIR, the knowledge distillation loss is used instead of classification loss. Usually, knowledge distillation is incorporated in transferring the knowledge from the teacher model to the student model. It is performed by minimizing the loss where the target is the result of the softmax function on the teacher model logits. However, the probability predicted by the model is usually very high for the true label and almost zero for the rest. Therefore, it doesn't provide additional information beyond ground truth has already provided. In order to resolve this issue, the \emph{softmax with temperature} was introduced \cite{hinton2015distilling}. The distillation loss is calculated as follows:
\begin{equation}
    L^D(x, \Tilde{y}; \theta) = -T^2 \sum_{c=1}^{N_{\text{classes}}}\Tilde{y}_c\log p_\theta^T(\mathcal{Y}=x|x),
\end{equation}
where T is the softmax temperature.

\subsection{Improved feature replay}

In this section, we describe three improvements that we propose to the base method that address particular problems with VAE-based feature replay: (1) reconstruction misalignment, (2) features drift in continual learning, (3) discrepancy between generated features and ones coming from the original data.

\subsubsection{Latent matching for reconstructions and original data}
\begin{figure}[htbp]
\centerline{\includegraphics[width=\linewidth]{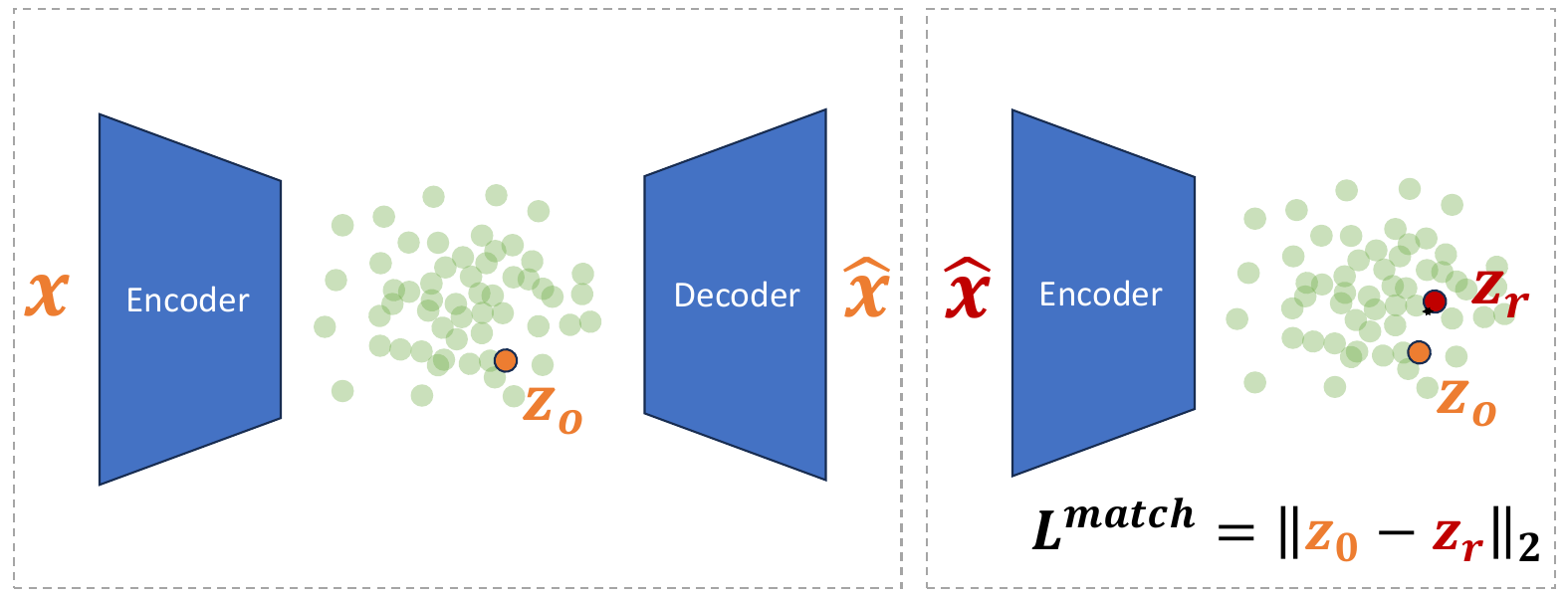}}
\caption{Visualisation of the latent matching loss. We minimize the difference between latent vectors of the original samples and their reconstructions.} \label{fig:latent_match}
\end{figure}

The first modification that we add aims to improve VAE model performance in continual retraining. To that end, we propose a latent matching regularisation that enforces encoder to reverse the decoding operation performed by the decoder.
In order to do that we pass the original sample $x$ through the encoder and obtain the latent vector for the original sample $z_o$. Then we pass this latent vector through the decoder to get the reconstruction $\hat{x}$. After that, we pass the reconstruction through the encoder again and receive the latent vector $z_r$. 
In particular, we calculate the regularisation on mean and variations outputted by the encoder. To that end, we utilize the mean squared error (MSE) loss for measuring the difference between two vectors. Therefore, we introduced latent match loss which is defined as the following:
\begin{equation}
    L^{\text{latent match}}(z_o; \phi,\psi) = -\frac{1}{2}(z_r-z_o)^2   
\end{equation}
The visualisation of our latent match loss is presented in Fig.~\ref{fig:latent_match}.

\subsubsection{Latent distillation}

\begin{figure}[htbp]
\centerline{\includegraphics[width=1\linewidth]{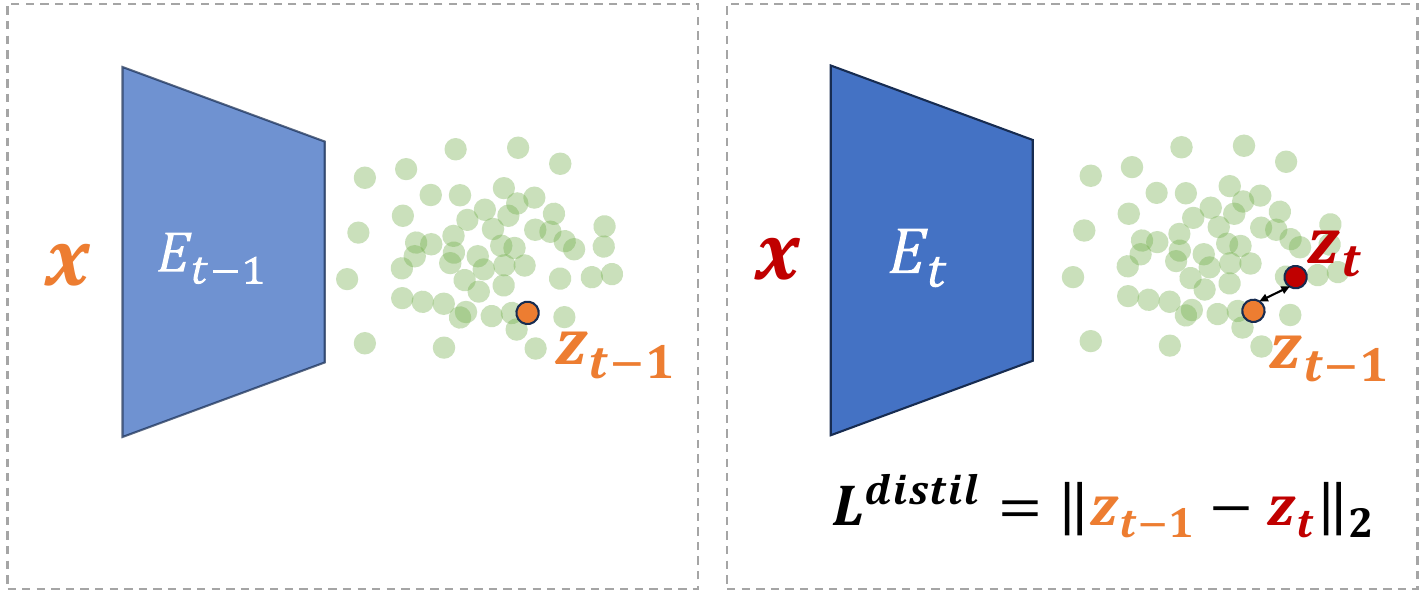}}
\caption{Visualisation of the latent distillation loss that reduces the feature drift between tasks.} \label{fig:latent_dist}
\end{figure}

\begin{figure*}[ht!]
\centerline{\includegraphics[width=0.7\linewidth]{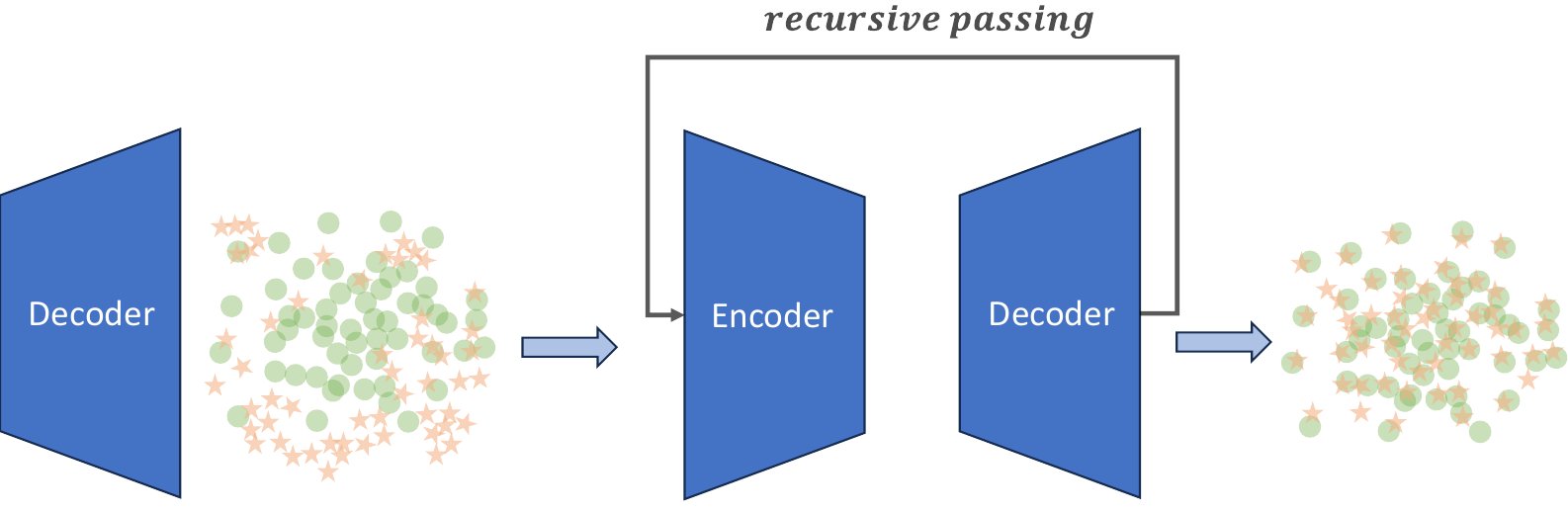}}
\caption{Visualisation of the cycling procedure. Each time we generate a batch of rehearsal samples (orange stars), we pass the generated outputs several times through the Variational Autoencoder in the recursive passing procedure. 
As a consequence, the final generations exhibit a considerably improved alignment with the reconstructions of the original training data (green dots).
} \label{fig:cycling_diag}
\end{figure*}

The BIR method, as described above in Sec~\ref{sec:base} has no mechanism for preventing feature drift, the change of distribution in features space over time as new tasks arrives. 
To prevent that, we add a latent distillation loss which is performed similarly as in~\cite{liu2020generative}. 
During the training of task $t$, we use the previously trained model consisting of encoder $E_{t-1}$ and decoder $D_{t-1}$. We use additional loss between the latent vector obtained by passing the sample through the previous model encoder $z_{t-1}$ and the latent vector produced by the current training model encoder $z_{t}$. The calculation of difference coincides with the calculation of latent matching loss defined before but with different inputs given:
\begin{equation}
    L^{\text{latent distill}}(z_{t-1}; \phi_{t-1,t},\psi_{t-1,t}) = -\frac{1}{2}(z_{t}-z_{t-1})^2
\end{equation}
The latent distillation loss serves as the purpose of the regularization term that controls forgetting, similarly to the SI regularization in the BIR method. Nevertheless our latent distillation achieves better performance.

\subsubsection{Cycling}

Even with the proposed changes, we hypothesize that there might be a significant difference between generated and original data features. To minimise this effect, we propose a cycling mechanism that is inspired by the idea presented by Gopalakrishnan et al. in~\cite{gopalakrishnan2022knowledge}. In this work, authors propose to recursively pass images from the buffer through the pre-trained autonecoder in order to better align them to the data from a new task. Here, we use the similar mechanism with our Variational Autoencoder to align generations of data from the previous task with data reconstructions. 

The visualisation of our cycling mechanism is presented in Fig.~\ref{fig:cycling_diag}.

\begin{figure}[ht]
\centerline{\includegraphics[width=0.9\linewidth]{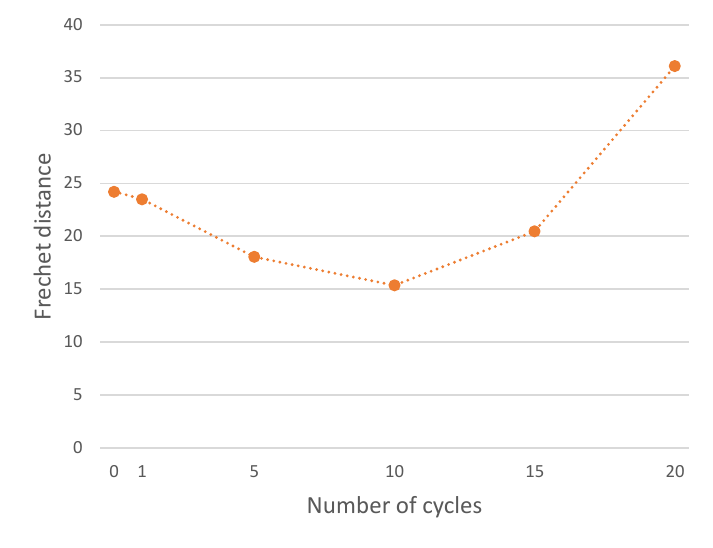}}
   \vspace*{-5mm}
 
\caption{Fr\'echet distance between original and generated latents as a function of a number of cycles. 0 stands for the standard model (no cycling). As we increase the number of cycles (up to some point) the generated latent vectors match more closely those from original data.
}
\label{comparisonfig}
\end{figure}

To verify our assumption we measure the distance between original features and generated ones we compute the Fr\'echet distance~\cite{fid}, which measures the distance between two Gaussian distributions. It is commonly used to compare the quality of generated images (also known as Fr\'echet inception distance), however, here we use it on the latent vector level. 
Figure~\ref{comparisonfig} shows how the Fr\'echet distance is reduced between generated latents and original ones as we use cycling. This motivates us to incorporate it during training.
Empirical evaluation of the cycling and number of used rounds is presented with other experiments in Sec.~\ref{sec:exp_cycl}.

\subsection{Final training objective}

To summarize, in our improved version of the baseline VAE method (BIR) we combine all of the described components into a single training objective function for the class-incremental learning session. It consists of two main parts namely $L^{\text{current}}$ and $L^{\text{replay}}$. $L^{\text{current}}$ is the loss that is calculated for the data of the current task, and it is given by:
\begin{equation}
    L^{\text{current}} = L^G + L^C + L^{\text{latent match}}
\end{equation}
$L^{\text{replay}}$ is calculated for the generations as follows:
\begin{equation}
    L^{\text{replay}} = L^G + L^D + L^{\text{latent distill}}
\end{equation}
The final loss function is the combination of these two losses:

\begin{equation}
    L^{\text{total}} = L^{\text{current}} + L^{\text{replay}}
\end{equation}

We use this loss to train the encoder, decoder, and classifier with current task data and data from the generative feature replay, additionally aligned with cycling through VAE. For the final loss, we start with a simple version without using any additional tradeoffs (coefficients) to balance each component. That can be further investigated. The ablation study is provided in Section~\ref{sec:ablation}. The steps of the overall training procedure can be found in the Algorithm \ref{alg:cap}.

\begin{algorithm}[t]
\small
\caption{Class-incremental learning with improved generative feature replay}\label{alg:cap}
\begin{algorithmic}
\renewcommand{\algorithmicensure}{\textbf{Input:}}
\Ensure Data $D_1$, $D_2$, ... , $D_T$, where $D_t = \{F(X_t), Y_t\}$, where F is a pretrained feature extractor
\Require Initialized encoder $Enc_0$, initialized decoder $Dec_0$, initialized classifier $\theta_{0}$, number of cycles $N_{cycles}$

\For{$t=1, \ldots ,T$}
\If{$t=1$}
    \State Step 1: Train $Enc_{new}$, $Dec_{new}$ and $\theta$ on data $D_1$ by\\\hspace{1.8cm} minimizing $L^{\text{current}}$
\Else
    \State Step 2: Save previously trained generator \\\hspace{1.8cm} $Dec_{old}=Dec_{new}$, $Enc_{old} = Enc_{new}$
    \State Step 3: Generate data $\hat{D}_{1:t-1}=Dec_{old}(y_{t'},z)$, \\\hspace{1.8cm} where $y_{t'}$ is all classes seen-so-far
    \State Step 4: 
    \hspace{1.8cm} \For{$k< N_{cycles}$} 
            \State $\hat{D}_{1:t-1}=Dec_{old}(Enc_{old}(\hat{D}_{1:t-1}))$ 
            \EndFor
    \State Step 5: Train $Enc_{new}$, $Dec_{new}$ and $\theta$ \\\hspace{1.8cm}on current data $D_t$ by minimizing $L^{\text{current}}$ \\\hspace{1.8cm}and on generated data $\hat{D}_{t-1}$ by minimizing $L^{\text{replay}}$
\EndIf
\EndFor
\end{algorithmic}
\end{algorithm}

\section{Experimental setup}

\subsection{Dataset}
We evaluate the models on two commonly used benchmarks that are challenging for the generative replay setup CIFAR-100 dataset~\cite{krizhevsky2009learning} and mini-ImageNet. CIFAR-100 consists of 100 object classes in 45,000 images for training, 5,000 for validation, and 10,000 for test. All images are in the size of 32$\times$32 pixels. The mini-ImageNet contains 50,000 training images, and 10,000 testing images evenly distributed across 100 classes. All images have the size 84$\times$84.

\subsection{Implementation details}
We utilize PyTorch as our framework \cite{paszke2017automatic}. For CIFAR-100 we use the ResNet-32 as the feature extractor pretrained on the first 50 classes of the dataset after randomly shuffling the data. For mini-ImageNet we extend the model to ResNet-18. For the pretraining stage, we use strong data augmentations from the PyCIL framework~\cite{zhou2023pycil}, which improves the feature extractor. In incremental steps, when we use an already pretrained feature extractor, we change data augmentation to one introducing less distortions to the inputs: images are firstly padded by 4 and then are randomly cropped to have size 32$\times$32 for CIFAR-100 and 84$\times84$ for mini-ImageNet. In addition, we use random horizontal flips for augmentation.
We train the encoder part on top of the feature extractor for 10000 iterations for the first task and for 5000 iterations for the rest of the tasks. Adam optimizer is used for the experiments with the learning rate equal to 1e-4. 

\subsection{Evaluation}
For evaluation, we use the average overall accuracy metric as in \cite{van2020brain}. It is the average accuracy of the model on the test data of all tasks up to the current one. In addition, to evaluate the overall performance, we calculate average incremental accuracy over all tasks. It is obtained by taking the average of accuracies after each task.
Each experiment is performed over 3 random seeds and the mean is reported.

\section{Results and Analysis}

\begin{table}
\begin{center}
{\caption{The average incremental accuracies on CIFAR-100 with the first task containing 50 classes and the rest 50 classes split into 5, 10, and 25 tasks equally}
\label{table1}}
\resizebox{0.9\columnwidth}{!}{%
\begin{tabular}{lccc}
\\[-6pt]
\textbf{CIL Method}&\textbf{T=6}&\textbf{T=11}&\textbf{T=26}\\
\hline
Finetune&32.41$\pm$0.07&23.42$\pm$0.09&13.26$\pm$0.16\\
SI&35.32$\pm$0.35&26.13$\pm$0.74&15.6$\pm$0.27\\
EWC&32.64$\pm$0.05&23.53$\pm$0.74&13.33$\pm$0.08\\
LwF&51.38$\pm$0.16&43.57$\pm$0.26&22.63$\pm$0.08\\
BIR&54.52$\pm$0.29&51.16$\pm$0.57&44.95$\pm$0.59\\
BIR+SI&57.18$\pm$0.23&52.4$\pm$0.29&47.71$\pm$0.98\\
Ours&\textbf{59.05$\pm$0.42}&\textbf{57.97$\pm$0.99}&\textbf{53.75$\pm$0.32}\\
\hline
\rule{0pt}{12pt}
Joint&\multicolumn{3}{c}{64.7}
\end{tabular}
} 
\end{center}
\end{table}

\begin{table}
\begin{center}
{\caption{The average incremental accuracies on mini-ImageNet with the first task containing 50 classes and the rest 50 classes split into 5, 10, and 25 tasks equally}
\label{table1_mini}}
\resizebox{0.9\columnwidth}{!}{%
\begin{tabular}{lccc}

\\[-6pt]
\textbf{CIL Method}&\textbf{T=6}&\textbf{T=11}&\textbf{T=26}\\
\hline
Finetune&29.9$\pm$0.08&22$\pm$0.05&13.04$\pm$0.03\\
SI&30.68$\pm$0.34&23.27$\pm$0.27&14.08$\pm$0.22\\
EWC&30.03$\pm$0.03&22.07$\pm$0.06&13.09$\pm$0.02\\
LwF&45.84$\pm$0.3&39.28$\pm$0.29&21.47$\pm$0.22\\
BIR&47.86$\pm$0.22&44.15$\pm$0.58&38.93$\pm$0.83\\
BIR+SI&49.59$\pm$0.85&47.52$\pm$0.4&43.78$\pm$0.55\\
Ours&\textbf{52.45$\pm$1.22}&\textbf{52.79$\pm$2.1}&\textbf{48.94$\pm$0.71}\\
\hline
\rule{0pt}{12pt}
Joint&\multicolumn{3}{c}{64.2}
\end{tabular}
} 
\end{center}
\end{table}

\begin{figure*}[htbp]
\centerline{\includegraphics[width=1\linewidth]{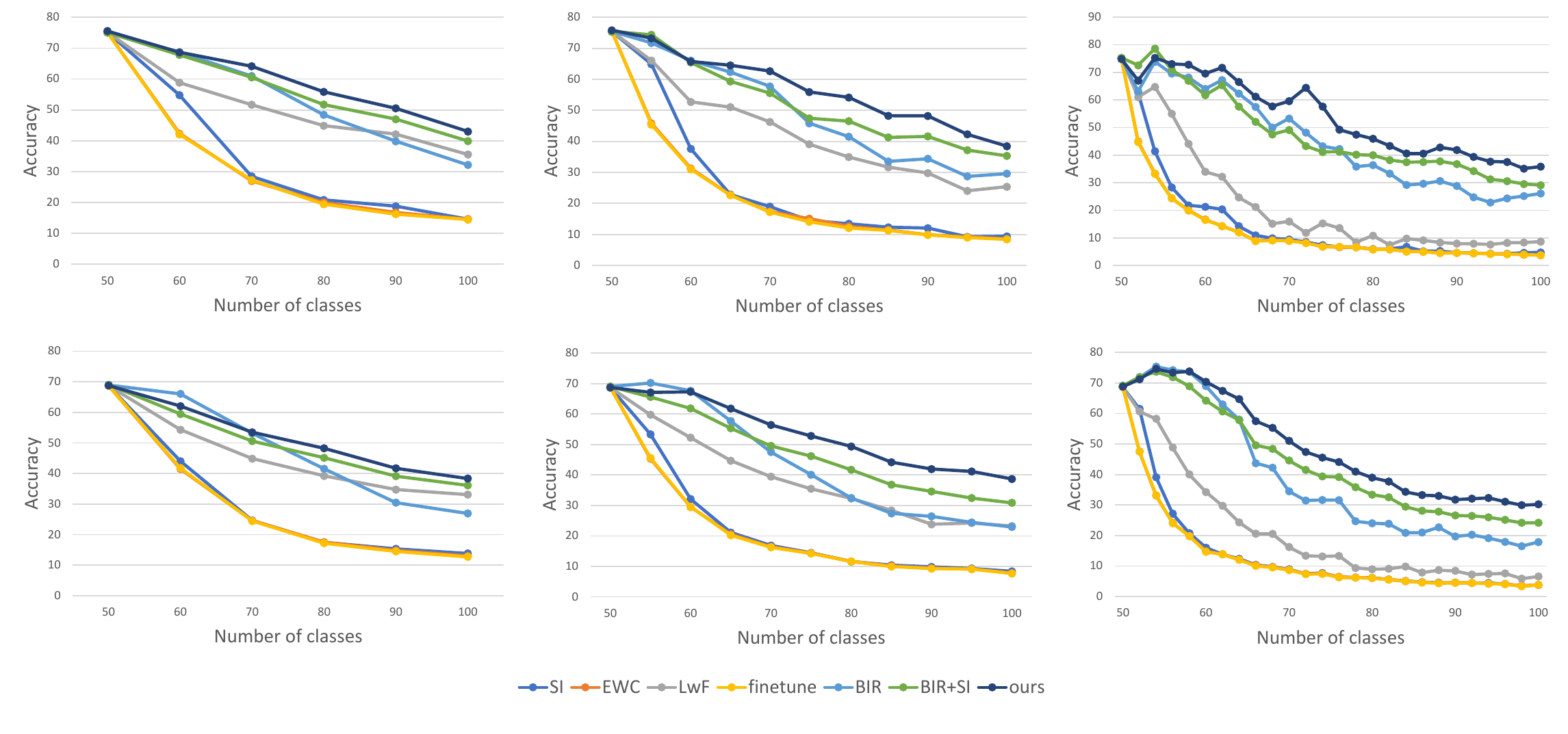}}
   \vspace*{-5mm}
\caption{Comparison of average accuracies on CIFAR-100 (top) and mini-ImageNet (bottom) after each task for 6, 11, and 26 tasks with the first task containing 50 classes} \label{accspic}
\end{figure*}

\subsection{Main results}

We performed the experiments on CIFAR-100 and mini-ImageNet with the first task containing 50 classes following \cite{van2020brain}. The rest 50 classes were split equally into 5, 10, and 25 tasks.

The average incremental accuracies for CIFAR-100 are shown in Table~\ref{table1}, and the accuracies after each task for T = 5, 10, 25 are shown in the form of plots in Figure~\ref{accspic} (top). Our method outperforms the regularization methods, and also the baseline BIR method. The second best method is BIR+SI, but, it is consistently worse than the proposed approach. 

Similar results are presented for mini-ImageNet dataset, which consists of bigger images than CIFAR-100. Table~\ref{table1_mini} present average incremental accuracy for this dataset. Here, as well for CIFAR-100, our method outperforms the other in a meaning of average incremental accuracy. However, the difference between ours and BIR+SI is more significant with the increasing number of tasks, where for T=26 we reach 48.94 and BIR+SI 43.78. The other regularization-based methods baselines for this scenario fall far behind. In Figure~\ref{accspic} (bottom) we see accuracies after each task. For mini-ImageNet BIR results in a better average accuracy in the second task for T=6 and T=11. This can be attributed to better plasticity (no SI). However, with a longer training and with more task, our method outperforms others.

For both datasets, SI alone presents the results comparable to finetuning. While simple application of LwF works good for smaller number of bigger tasks, T=6 and T=11, but for longer sessions T=26 the performance significantly drops. Here, better adjustment of regularization hyper-parameters can play more important role. Our proposed method does not suffer from this issue. 

\subsection{Number of cycles}
\label{sec:exp_cycl}
\begin{figure}[htp]
\centerline{\includegraphics[width=0.95\linewidth]{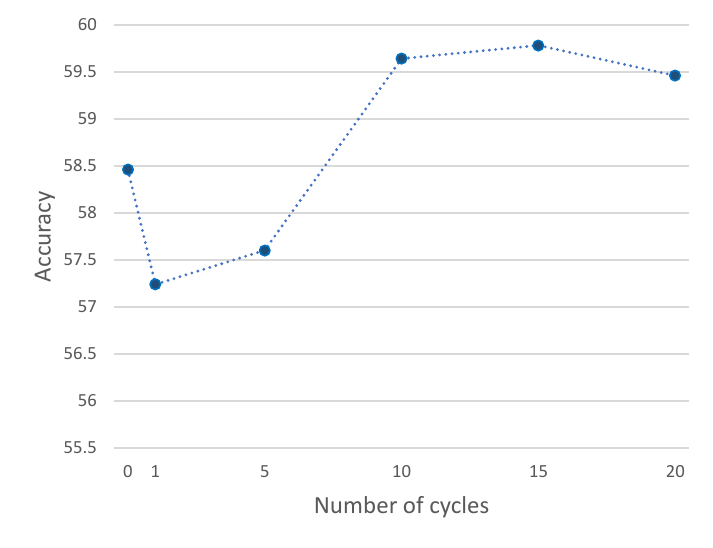}}
   \vspace*{-5mm}

\caption{Average incremental accuracy of the model depending on number of cycles for \textit{T=6}} \label{cyclesfig}
\end{figure}

We perform the analysis of how the number of cycles influences the average incremental accuracy for $T=6$. Figure~\ref{cyclesfig} shows the accuracy firstly drops but with an increased number of cycles the performance improves significantly. The number of cycles should be treated as a hyperparameter and tuned for different datasets and split scenarios.

\subsection{Ablation study}
\label{sec:ablation}

We perform an ablation study of our method. By starting from the baseline model (BIR), we add one by one the modifications that we propose. The results of the ablations study are presented in Table~\ref{tab:ablation}. As can be seen, all the elements of our method contribute significantly to the overall performance, where in total we reach $5.56\%$ of average incremental accuracy in comparison to BIR.

\begin{table}
\centering
\caption{Ablation study of our method for class incremental learning setting with T=6 and CIFAR-100. Average incremental accuracy is reported for ResNet32.}
\label{tab:ablation}
\resizebox{\columnwidth}{!}{%
\begin{tabular}{lcccc} 
 \hline

 Approach & Latent match & Latent  & 10 cycles & Acc.(\%) \\ 
 & & distillation & & \\
 \hline 
baseline method - BIR & & & & 54.22\\ 
\hline

w/ latent match &\checkmark& & &56.21\\
w/ latent distillation& \checkmark&\checkmark& &  58.46\\ 
w/ 10 cycles & \checkmark&\checkmark& \checkmark& 59.78\\
\hline
\end{tabular}
}
\end{table}

\begin{figure*}[ht]
    \centering
    \begin{subfigure}[b]{0.45\textwidth}
        \centering
        \includegraphics[width=\textwidth]{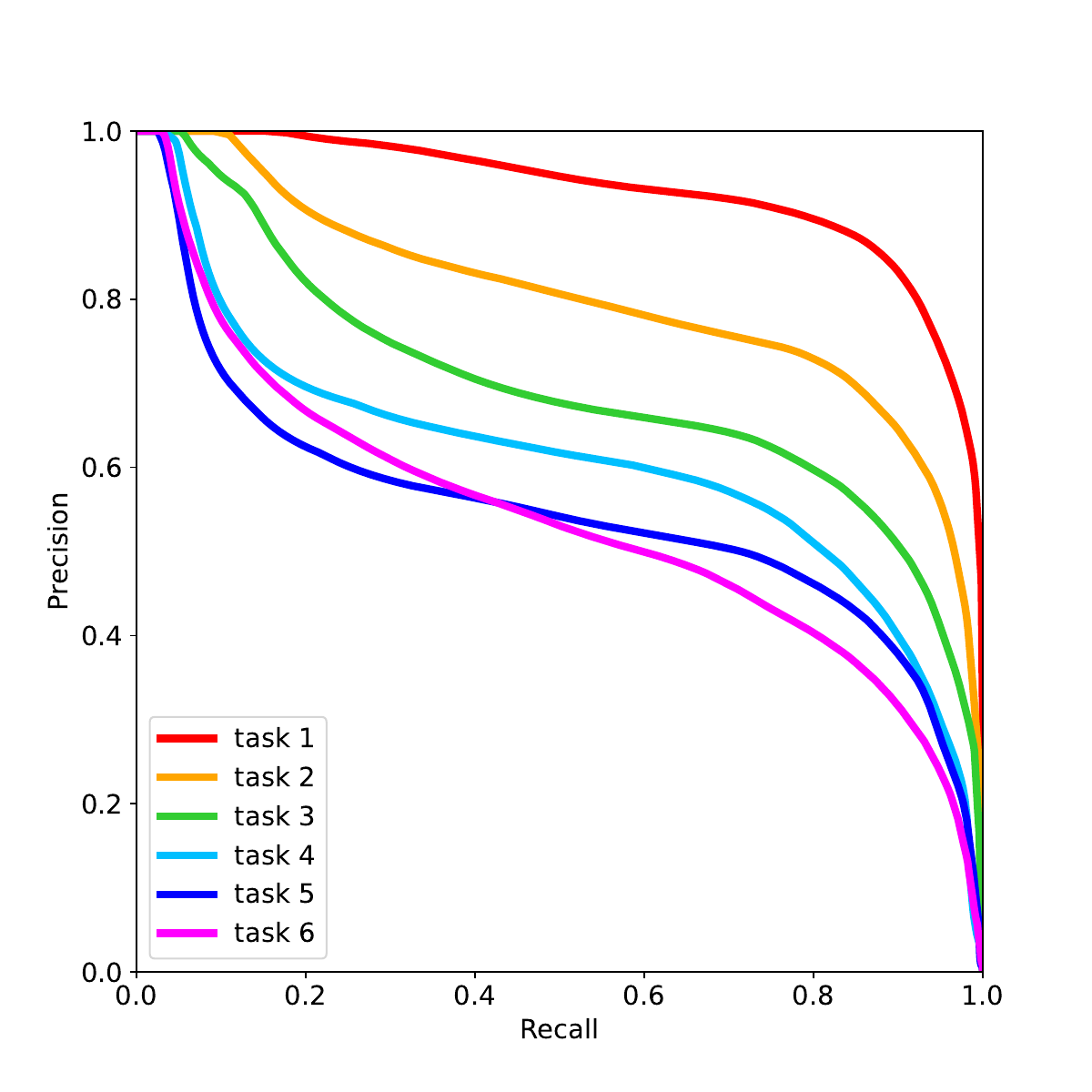}
    \end{subfigure}
    \begin{subfigure}[b]{0.45\textwidth}
        \centering
        \includegraphics[width=\textwidth]{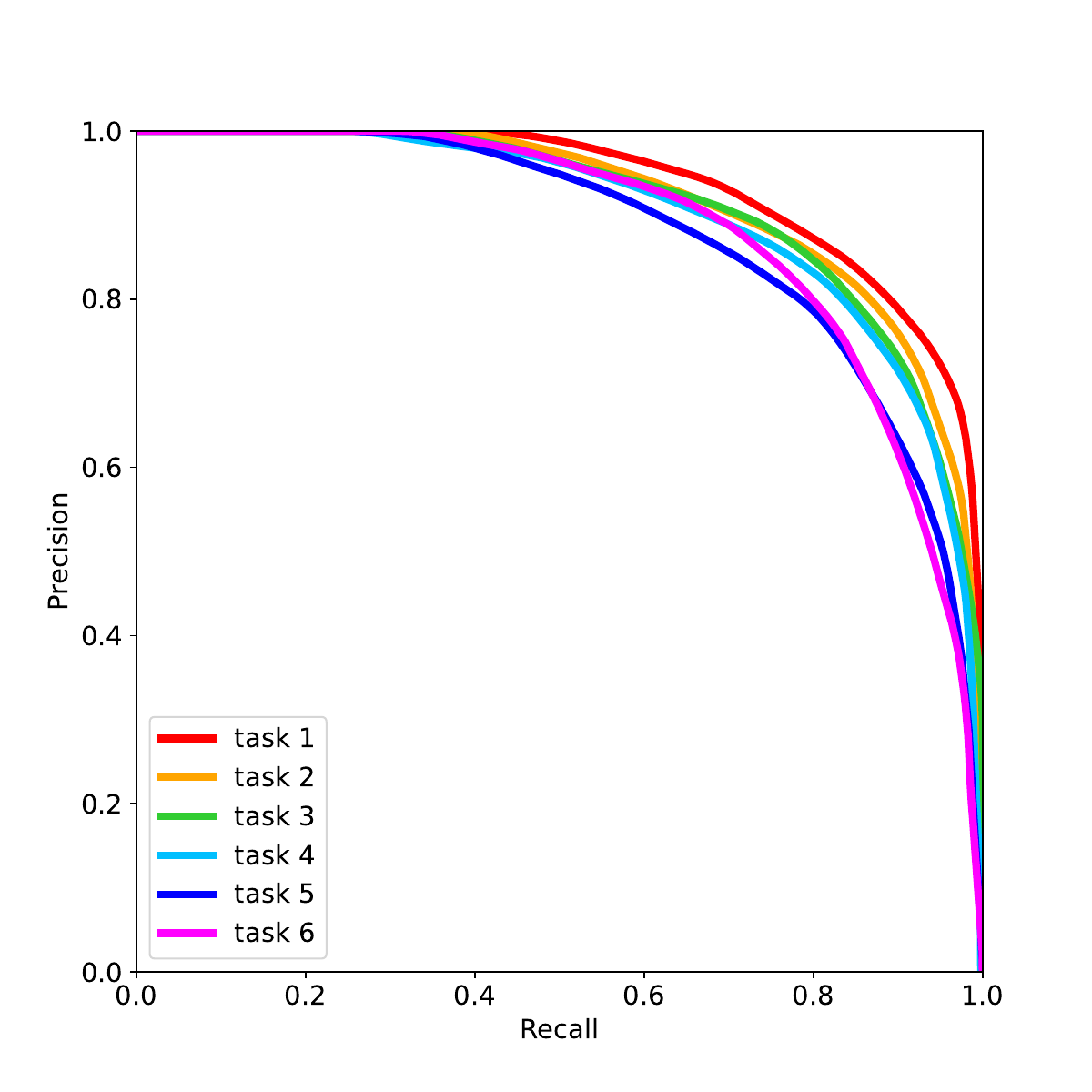}
    \end{subfigure}
    \vspace*{-5mm}
   \caption{
  Comparison of the Precision/Recall curves for features generated after each task with either the standard BIR method (left) or our improved version (right). Our method is able to retain much better precision-recall tradeoff of the generated samples.}
   \label{fig:prec_rec}
\end{figure*}

\subsection{Analysis of Precision and Recall}

Finally, we perform the analysis of our models performance in terms of the quality of generations. To that end, we refer to the distribution precision and recall of the distributions as proposed by~\cite{sajjadi2018assessing}. As authors indicate, those metrics disentangle FID score into two aspects: the quality of generated results (Precision) and their diversity (Recall). We calculate those two metrics on the features level and compare the resulting scores between standard BIR method and our improved approach. As presented in Figure~\ref{fig:prec_rec}, our improvements allow the model to retain both higher precision and recall of the regenerated samples. 

\section{Conclusions and Future Work}
In this work, we propose a set of improvements for generative replay in class incremental learning. We observe that the currently used approach for feature-level replay suffers from the mismatch of latent vectors between original and regenerated samples. Based on that we add a loss function that aligns the latent vectors together. On top of that, we have proposed a cycling procedure, which passes the generated features through the model several times, before being used in the training. This allowed us to scale the generative approaches to more complex datasets, such as mini-ImageNet. Through, the ablation study we have shown the improvements coming from each of the introduced components.

For future work, we aim to scale the proposed solution to more challenging datasets, such as ImageNet, and longer sequences of more diversified tasks. This stands out as a notable limitation in numerous generative replay methods which are unsuitable for larger datasets, whereas our approach holds a significant advantage in this regard. Another interesting future work direction is to prepare VAE-based feature replay models for task-free scenarios in CIL.

\paragraph{Impact Statement.} 
By using the generative approach for continual learning, our method does not require storing exemplars of past data, therefore it addresses concerns about private or sensitive data, which are applicable in some scenarios. However, generative models can retain the biases present in the training data, and we strongly advise a careful examination of their performance to ensure unbiased outcomes.

{\small
\bibliographystyle{ieee_fullname}
\bibliography{egbib}
}

\end{document}